\documentclass[conference]{IEEEtran}
\ifCLASSINFOpdf
\else
\fi

\usepackage{url}
\usepackage{graphicx}

\newcommand{\midtilde}{\raisebox{-1mm}{\textasciitilde}}

\begin{document}
%
\title{ImageNet MPEG-7 Visual Descriptors \\ \huge{Technical Report}}

\author{\IEEEauthorblockN{Fr\'{e}d\'{e}ric Rayar
\IEEEauthorblockA{Universit\'{e} Fran\c{c}ois Rabelais of Tours,\\ LI EA-6300, France\\
frederic.rayar@univ-tours.fr}}
}


%


\maketitle

\begin{abstract}
ImageNet is a large scale and publicly available image database. It currently offers more than 14 millions of images, organised according to the WordNet hierarchy. One of the main objective of the creators is to provide to the research community a relevant database for visual recognition applications such as object recognition, image classification or object localisation. However, only a few visual descriptors of the images are available to be used  by the researchers. Only SIFT-based features have been extracted from a subset of the collection. This technical report presents the extraction of some MPEG-7 visual descriptors from the ImageNet database. These descriptors are made publicly available in an effort towards open research.

\end{abstract}


%
\IEEEpeerreviewmaketitle

\section{Introduction}

ImageNet~\cite{Deng:2009} is a large scale image ontology database. The investigators of this project, namely Pr. Li Fei-Fei and her colleagues at Princeton University, have started their efforts in 2009. They believe that such a database is \textit{``a critical resource for developing advanced, large-scale content-based image search and image understanding algorithms, as well as for providing critical training and benchmarking data for such algorithms.''} Such a resource can be useful in visual recognition applications such as object recognition, image classification or object localisation.\\

ImageNet is organised according to the well-known WordNet~\cite{Fellbaum:1998} ontology. It is a \textit{``large lexical database of English. Nouns, verbs, adjectives and adverbs are grouped into sets of cognitive synonyms (synsets), each expressing a distinct concept''}. Hence, ImageNet can be viewed as a tree where each node corresponds to a synset. Each node contains 500 to 1000 images to illustrate the associated synset. Images of each concept have been quality-controlled and human-annotated. More details on the curation process can be found in the creators paper~\cite{Deng:2009}. The database and related resources are publicly available at \url{http://www.image-net.org}. Currently, ImageNet has 14,197,122 images and 21,841 synsets indexed. In addition to the images, one can find SIFT~\cite{Lowe:2004} based features or object attributes, but only for images of small subsets of the 21,841 synsets (1000~\cite{ImageNet:sbow} and 384~\cite{ImageNet:oa} synsets respectively).\\

In this work, we aim at providing to the community two MPEG-7 visual descriptors for the whole ImageNet dataset. The rest of the paper is organises as follows: the MPEG-7 visual description standard is succinctly presented in Section~\ref{sec:mpeg}. The two computed descriptors, namely Colour Layout Descriptor (CLD) and Edge Histogram Descriptor (EHD), are described in subsection~\ref{sec:cld} and~\ref{sec:ehd} respectively. Finally, the provided data sets are presented in Section~\ref{sec:data}.\\

\section{MPEG-7}
\label{sec:mpeg}
Following its previous MPEG version, where the emphasis was put on encoding, MPEG-7~\cite{Manjunath:2002} aims at describing multimedia content, such as image, audio and video.
MPEG-7 specifies, among others, visual descriptors that have been introduced and stressed out in the research community. The objective is to provide a set of
standardised descriptors to characterise multimedia content in order to guaranty interoperability and achieve content based multimedia retrieval efficiently. \\

These visual descriptors~\cite{Sikora:2001} are organised in four categories:
\begin{itemize}
	\item colour descriptors~\cite{Manjunath:2001}
	\item texture descriptors~\cite{Manjunath:2001}
	\item shape descriptors~\cite{Bober:2001}
	\item motion descriptors (for videos)~\cite{Jeannin:2001}
\end{itemize}
One can refer to the references attached to each category to find a detailed description of each descriptors.\\

We describe below the two descriptors that have been computed in this work, namely Colour Layout Descriptor and Edge Histogram Descriptor (colour and texture descriptor respectively). Note that the original definition are presented in this paper. One can find some studies that have improve the basic definition of CLD and EHD (\textit{e.g.} in~\cite{ehd:2002}).\\

\subsection{Colour Layout Descriptor}
\label{sec:cld}
The CLD is designed to capture the spatial distribution of dominant
colours in an image. The colours are expressed in the YCbCr colour space. CLD main advantages are first that it is a very compact descriptor, therefore it fits perfectly for fast browsing and search applications. Second, it is resolution invariant, thanks to its definition.\\

\begin{figure}[!ht]
\center
	\includegraphics[width=0.9\linewidth]{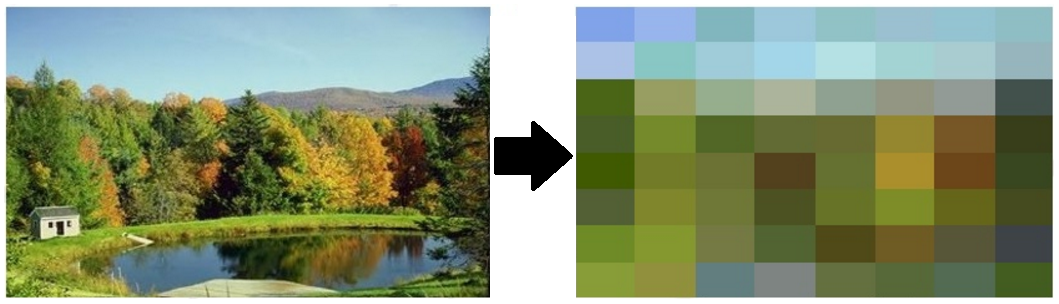}
	\caption{CLD: representative colour detection on a $8\times8$ partitioned image.}
	\label{fig:colorgrid}
\end{figure}

\begin{figure}[!ht]
\center
	\includegraphics[width=0.4\linewidth]{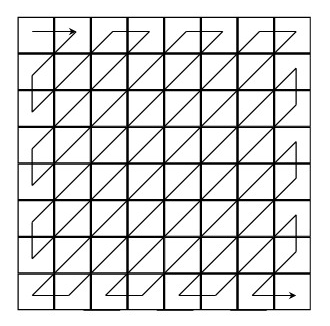}
	\caption{CLD: zig-zag re-ordering.}
	\label{fig:zigzag}
\end{figure}

The feature extraction process consists in four stages:
\begin{itemize}
	\item \textit{image partitioning}: the input picture is divided into \\$8\times8=64$ blocks. This guarantee the resolution or scale invariance.
	\item \textit{representative colour detection}: one representative colour is computed for each block. To select the representative colour, any method can be applied, but 
the average colour of the block is usually used. An $8\times8$ icon is obtained (see Figure~\ref{fig:colorgrid}).
	\item \textit{DCT transformation}: each colour channel of the icon is transformed into a series of coefficients by performing a $8\times8$ Discrete Cosine Transform (DCT). Three series of 64 DCT coefficients are computed.
	\item \textit{quantization}: a few low-frequency coefficients are selected using zigzag scanning (see Figure~\ref{fig:zigzag}) and quantised to form the descriptor.\\
\end{itemize}

Hence, we obtained a descriptor with a size of $3\times64=192$. One can find more details on the the CLD computation in~\cite{Manjunath:2001} or~\cite{Manjunath:2002}.

\subsection{Edge Histogram Descriptor}
\label{sec:ehd}
\begin{figure}[!ht]
\center
	\includegraphics[width=0.8\linewidth]{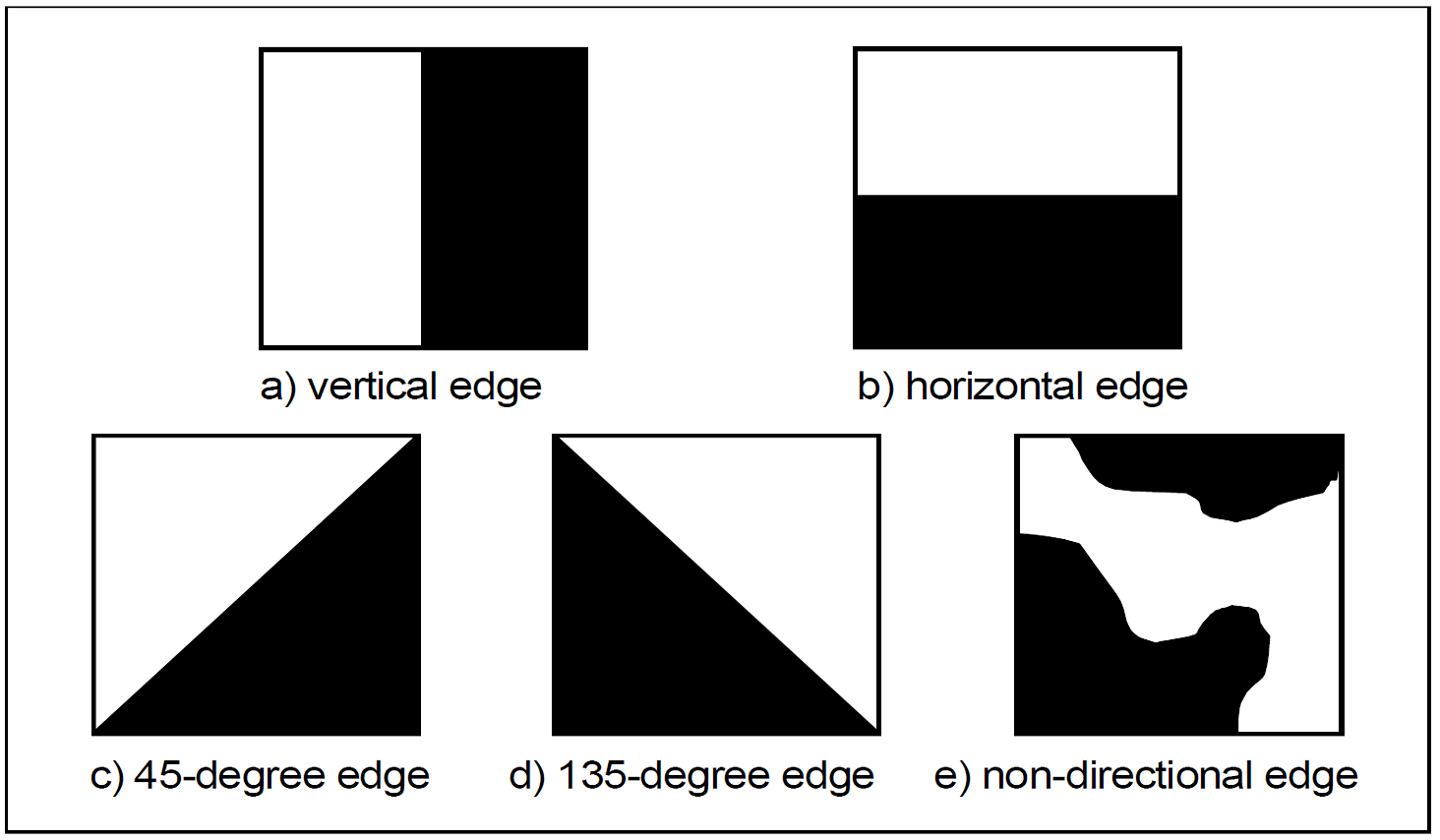}
	\caption{EHD: 5 considered types of contour. Illustration from \cite{ehd:2002}.}
	\label{fig:edges}
\end{figure}

EHD~\cite{ehd:2000} captures the spatial distribution of five types of edges in an image. In that sense, CLD and EHD can be viewed as alike as two peas in a pod. Figure~\ref{fig:edges} shows the five edges considered.\\

\begin{figure}[!ht]
\center
	\includegraphics[width=0.8\linewidth]{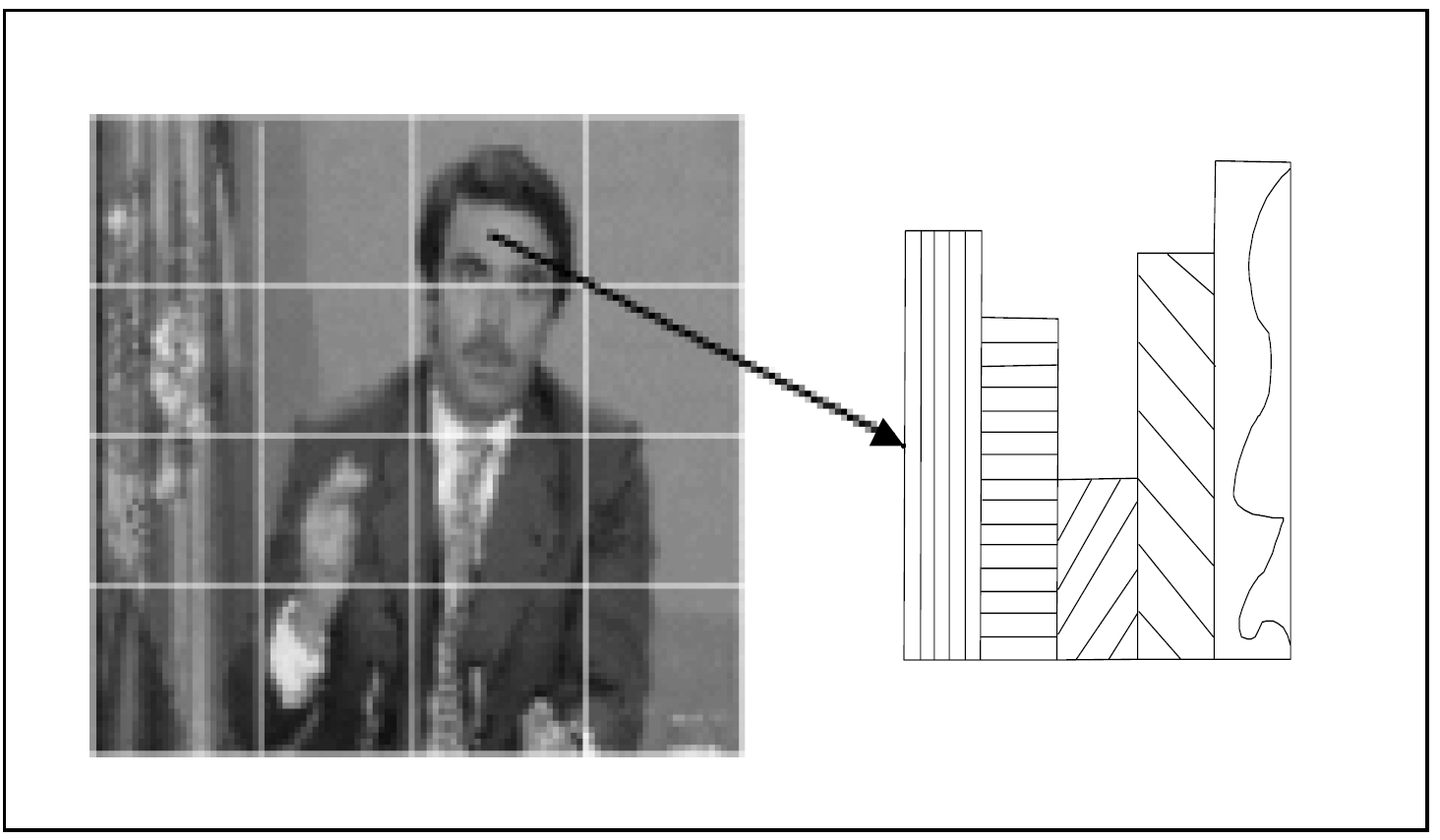}
	\caption{EHD: histogram computation for each cell. Illustration from \cite{ehd:2002}.}
	\label{fig:zoning}
\end{figure}

The EHD feature extraction is quite straightforward: first, an image is divided using a $4\times4$ blocks. For each block, an histogram of the distribution of the considered edges is built (see Figure~\ref{fig:zoning}). Therefore, we have $4\times4\times5=80$ values that constitutes the texture descriptor.\\

One can find more details on the the EHD computation in~\cite{Manjunath:2001}.
	
\section{Files Description}
\label{sec:data}

\subsection{Database}
\begin{itemize}
	\item All the 21,841 synsets have been processed
	\item A total of 14,197,060 images have been processed.
	\item The remaining 62 images have not processed. They  mostly correspond to corrupted image files.
\end{itemize}

\subsection{Files format}
\begin{itemize}
	\item There is one file per synset, named \textit{synset\_id.ext}\\ with ext = \{cld, ehd\}.
	\item Each file contains the descriptors for the $n$ images associated with its synset. 
	\item Hence, for a given synset, the associated file contains $n$ lines with the scheme
\begin{verbatim}
	image_id;value_1;value_2;...;value_d;
\end{verbatim}	
with 	
\[
	d = \left\{
		\begin{array}{rc}
			192 & \textrm{if CLD} \\
			 80 & \textrm{if EHD}
		\end{array}
	\right.
\]
\end{itemize}

\subsection{Files availability}
\begin{itemize}
	\item The following archives have been made publicly available:
		\begin{itemize}
			\item \textbf{ImageNet\_MPEG-7\_CLD\_whole.zip}~\footnote{\url{http://rfai.li.univ-tours.fr/PublicData/ImageNetFeatures/ImageNet_MPEG-7_CLD_whole.zip}} (\midtilde 1GB)
			\item \textbf{ImageNet\_MPEG-7\_EHD\_whole.zip}~\footnote{\url{http://rfai.li.univ-tours.fr/PublicData/ImageNetFeatures/ImageNet_MPEG-7_EHD_whole.zip}} (\midtilde 2GB)
		\end{itemize}
		\vspace{1mm}
	\item \textit{Note}: some images \textit{might} not have both CLD and EHD. However, it should not be the case. We just did not checked this.
\end{itemize}

\bibliographystyle{IEEEtran}
\bibliography{ImageNET}

\end{document}